\documentclass[sigconf, nonacm]{acmart}
\usepackage{array}
\usepackage{adjustbox}
\usepackage{subcaption}
\AtBeginDocument{%
  }

\begin{document}

\title{UniDGF: A Unified Detection-to-Generation Framework for Hierarchical Object Visual Recognition}

\settopmatter{authorsperrow=4}

\author{Xinyu Nan}
\authornote{Co-first authors.}
\affiliation{%
  \institution{Kuaishou Technology}
  \city{Beijing}
  \country{China}
}
\email{2301213095@stu.pku.edu.cn}

\author{Lingtao Mao}
\authornotemark[1]
\affiliation{%
  \institution{Kuaishou Technology}
  \city{Beijing}
  \country{China}
}
\email{mltzju@163.com}

\author{Huangyu Dai}
\authornotemark[1]
\affiliation{%
  \institution{Kuaishou Technology}
  \city{Beijing}
  \country{China}
}
\email{11931034@zju.edu.cn}

\author{Zexin Zheng}
\affiliation{%
  \institution{Kuaishou Technology}
  \city{Beijing}
  \country{China}
}
\email{zhengzx25@mail2.sysu.edu.cn}

\author{Xinyu Sun}
\affiliation{%
  \institution{Kuaishou Technology}
  \city{Beijing}
  \country{China}
}
\email{sxy001122@gmail.com}

\author{Zihan Liang}
\affiliation{%
  \institution{Kuaishou Technology}
  \city{Beijing}
  \country{China}
}
\email{liangzih@seas.upenn.edu}

\author{Ben Chen}
\authornote{Corresponding author.}
\affiliation{%
  \institution{Kuaishou Technology}
  \city{Beijing}
  \country{China}
}
\email{benchen4395@gmail.com}

\author{Yuqing Ding}
\affiliation{%
  \institution{Kuaishou Technology}
  \city{Beijing}
  \country{China}
}
\email{kokia.ding@gmail.com}

\author{Chenyi Lei}
\affiliation{%
  \institution{Kuaishou Technology}
  \city{Beijing}
  \country{China}
}
\email{leichy@mail.ustc.edu.cn}

\author{Wenwu Ou}
\affiliation{%
  \institution{Kuaishou Technology}
  \city{Beijing}
  \country{China}
}
\email{ouwenweu@gmail.com}

\author{Han Li}
\affiliation{%
  \institution{Kuaishou Technology}
  \city{Beijing}
  \country{China}
}
\email{lihan08@kuaishou.com}

\renewcommand{\shortauthors}{Xinyu et al.}

\begin{abstract}
Achieving visual semantic understanding requires a unified framework that simultaneously handles object detection, category prediction, and attribute recognition. However, current advanced approaches rely on global similarity and struggle to capture fine-grained category distinctions and category-specific attribute diversity, especially in large-scale e-commerce scenarios. To overcome these challenges, we introduce a detection-guided generative framework that predicts hierarchical category and attribute tokens. For each detected object, we extract refined ROI-level features and employ a BART-based generator to produce semantic tokens in a coarse-to-fine sequence covering category hierarchies and property–value pairs, with support for property-conditioned attribute recognition. Experiments on both large-scale proprietary e-commerce datasets and open-source datasets demonstrate that our approach significantly outperforms existing similarity-based pipelines and multi-stage classification systems, achieving stronger fine-grained recognition and more coherent unified inference.
\end{abstract}

\ccsdesc[500]{Computing methodologies~Object detection}
\ccsdesc[500]{Computing methodologies~Object recognition}
\ccsdesc[500]{Computing methodologies~Object identification}
\ccsdesc[300]{Computing methodologies~Scene understanding}

\keywords{Generative Visual Recognition,
Semantic Hierarchical Encoding,
Object Detection,
Category Prediction,
Attribute Recognition}

\maketitle

\section{Introduction}
Achieving accurate and efficient semantic understanding of all objects in an image has long been a central goal in computer vision. Although recent advances in object detection and multimodal perception have made significant progress in their respective domains, most existing pipelines still rely on cascaded modules that lack integrated cross-level interaction~\cite{khanam2024yolov11,Cheng2024YOLOWorld,chen2023ovarnet}. Such disconnected designs often lead to suboptimal performance.

Recent representative efforts such as YOLO-World~\cite{Cheng2024YOLOWorld} and OvarNet~\cite{chen2023ovarnet} integrate object detection models (e.g., YOLO or RPN-based pipelines) with vision–language models (e.g., CLIP~\cite{radford2021learning}) to jointly handle detection, classification, and attribute recognition within a single framework. YOLO-World extends the YOLO architecture by leveraging CLIP-derived textual representations and feature alignment to enable open-vocabulary detection, while OvarNet~\cite{chen2023ovarnet} adds an attribute branch to predict semantic categories and visual attributes simultaneously. Although these approaches represent meaningful progress toward multi-task integration, their semantic outputs remain coarse and lack cross-level interaction, hindering coherent coarse-to-fine reasoning. Consequently, they often fail to capture fine-grained category distinctions and category-specific attribute diversity, especially in large-scale e-commerce scenarios with broad category coverage and diverse attribute spaces.

\begin{figure*}[t]
    \centering
    \includegraphics[width=\linewidth]{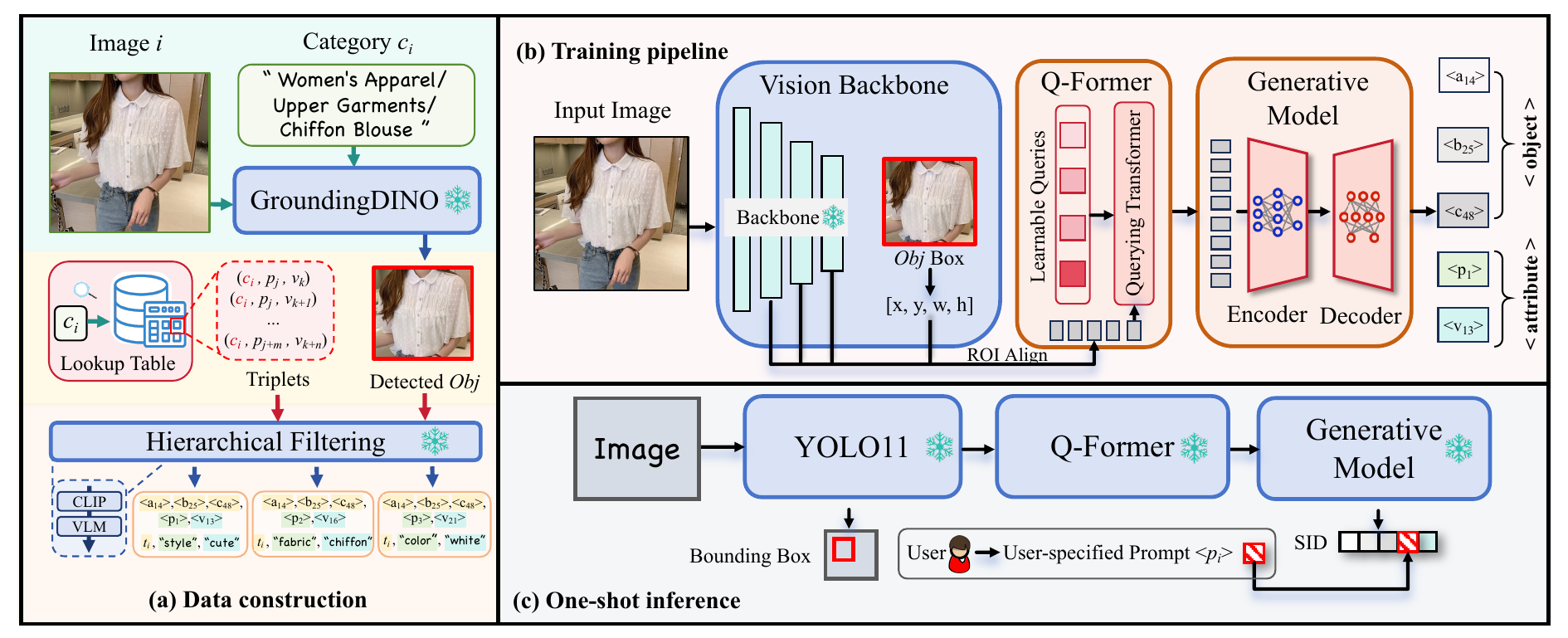}  
    \caption{Overview of the proposed UniDGF framework, including (a) Data construction, (b) Training pipeline, and (c) One-shot inference pipeline.}
    \label{fig:fig_main}
\end{figure*}
\begin{figure}
    \centering
    \includegraphics[width=\linewidth]{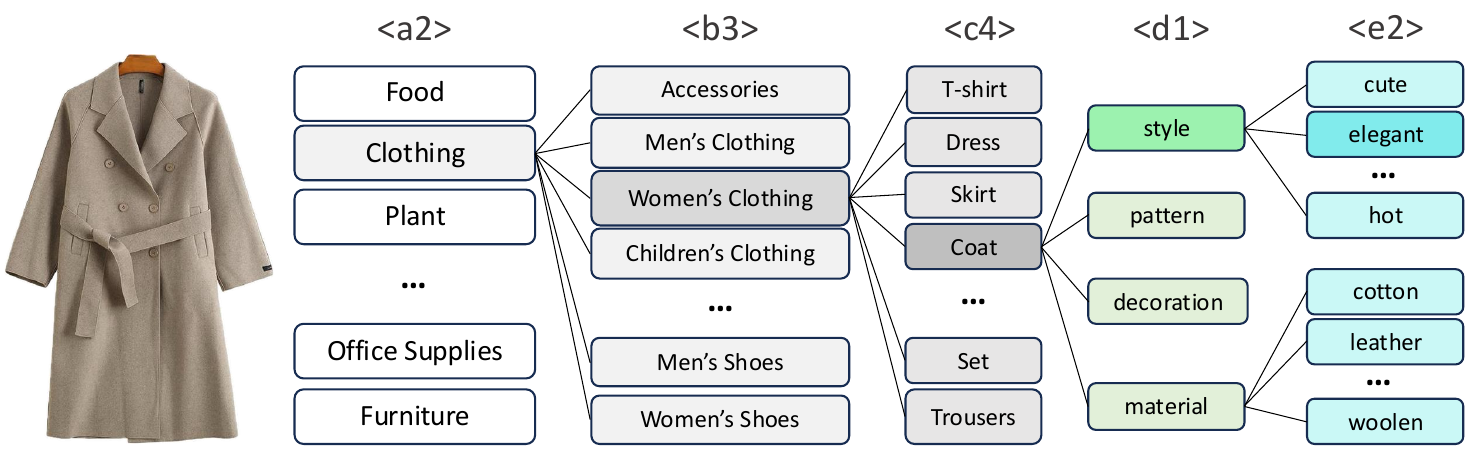}  
    \caption{Illustration of one data sample of hierarchical semantic encoding.}
    \label{fig:fig_hie}
    \vspace{-2em}
\end{figure}
To address these limitations, we propose a unified framework that combines a YOLO-based object detector with a generative hierarchical model for predicting object categories and diverse property–value pairs. The detector first identifies and localizes all items in the image. The extracted region features are then processed and fed into a BART generator, which outputs category and attribute IDs in a coarse-to-fine sequence. This design enables fine-grained semantic modeling that existing cascaded detect-then-recognize pipelines cannot effectively represent. In addition, our framework supports property-conditioned attribute recognition, allowing precise extraction of specific attributes such as color, material, or style. To mitigate the shortage of large-scale fine-grained annotations, we introduce a two-stage labeling pipeline in which CLIP first performs coarse filtering over candidate labels, and a vision–language model further refines the filtered candidates into high-quality annotations, from which we construct the large-scale Products7417 dataset. Extensive experiments on real e-commerce datasets and open-source datasets show that our approach consistently outperforms a wide range of paradigms, including direct classification models, semantic matching–based retrieval systems, and multimodal large language models, demonstrating the effectiveness of generative hierarchical prediction in large-vocabulary scenarios. We will release all validation data containing images and hierarchically encoded category and attribute labels, along with the pretrained weights of our model.

\section{Methodology}
\subsection{Dataset Collection Process}

Fig.~\ref{fig:fig_main}(a) summarizes the pipeline used to construct the Products7417 dataset.
(1) Approximately 1 million high-quality e-commerce product images are collected, each accompanied by its catalog-level category annotation.
(2) Leveraging the catalog category as prior knowledge, GroundingDINO~\cite{liu2024grounding} is applied to localize the primary product instance in each image to extract corresponding bounding box.
(3) A hierarchical category–attribute tree is then built to organize product categories, attribute types, and their associated values.
This hierarchy enables reversible coarse-to-fine semantic decomposition (e.g., \textit{"Women's Apparel"} ($\langle a_{14}\rangle$) $\rightarrow$ \textit{"Upper Garments"} ($\langle b_{25}\rangle$) $\rightarrow$ \textit{"Chiffon Blouse"} ($\langle c_{48}\rangle$) $\rightarrow$ \textit{"style"} ($\langle p_{1}\rangle$) $\rightarrow$ \textit{"cute"} ($\langle v_{13}\rangle$)).
(4) For each product, all candidate attribute–value pairs are retrieved from the corresponding branch of the hierarchy. CLIP is used to rank these candidates, proposing the top 3 attribute types and top 10 values, while Qwen2.5-VL-72B subsequently verifies which attribute–value pairs faithfully reflect the object's visual appearance.


In summary, each dataset sample contains the raw image, catalog category, detected bounding box, and the final hierarchical category–attribute–value triplet $(c_{i}, p_{j}, v_{k})$. 
Using the same pipeline, we also complete the missing category and attribute annotations for MSCOCO and Objects365.

\begin{table*}[t]
\centering
\caption{Comparison of object category and attributes prediction accuracy (\%) on MSCOCO, Objects365 and Products7417.}
\vspace{-0.2cm}
\begin{adjustbox}{max width=0.97\textwidth}
\begin{tabular}{ >{\centering\arraybackslash} p{2.5cm} >{\centering\arraybackslash}p{2.1cm} >{\centering\arraybackslash}p{3.0cm} cc cc cc}
\hline
\multicolumn{3}{c}{\textbf{Model}} &\multicolumn{2}{c}{\textbf{MSCOCO~\cite{lin2014microsoft,yun2022attributes}}}&\multicolumn{2}{c}{\textbf{Objects365~\cite{shao2019objects365}}}&\multicolumn{2}{c}{\textbf{Products7417}}\\
 \hline
 model name & vision backbone & inference strategy & Cate Acc. & Attr Acc. & Cate Acc. & Attr Acc. & Cate Acc. & Attr Acc. \\
\hline
YOLO-World & YOLO & embedding retrieval & 71.16	& -	& 48.68	& - & 12.93 & - \\ 
YOLOv11 & YOLO & classifier-based pred. & 82.18	& -	& 66.79	& - & 43.58	& - \\
UniDGF & YOLO & encoder–decoder gen. & 95.22 & \textbf{52.10} &	81.26	& \textbf{75.45} & \textbf{61.39} & \textbf{31.00}\\ 
UniDGF & YOLO & decoder-only gen. & \textbf{96.02} & 16.04 & \textbf{84.81} & 36.13 & 47.03 &  26.90\\ 
\hline
CLIP L/14 & ViT & embedding retrieval & 48.00 & 15.99 & 28.40 & 34.45 & 13.60 & 1.04\\ 
GME-Qwen2VL-7B & ViT & embedding retrieval & 62.96	& 21.74	& 33.43	& 40.21	& 23.61	& 1.16\\ 
Qwen2.5-VL-3B & ViT & prompt-based pred. & 44.49 & 13.67 & 29.90 & 38.71 & 18.75 & 6.10\\ 
DINOv3 & ViT & classifier-based pred. & 87.54 & 36.84 & 72.39 & 63.62 & 48.87 & 28.58
\\ 
UniDGF & ViT & encoder–decoder gen. & 91.02 & \textbf{50.94} & \textbf{79.00} & \textbf{71.11} & \textbf{51.81} & \textbf{30.92}
\\ 
UniDGF & ViT & decoder-only gen. & \textbf{91.48}	& 15.74	& 77.49	& 37.43	& 49.68	& 26.72\\ 
\hline
\end{tabular}
\end{adjustbox}
\label{table_ACC}
\end{table*}

\begin{table}
\centering
\caption{Comparison of YOLO-based object detectors in terms of mAP(\%) on MSCOCO, Objects365 and Products7417.}
\begin{adjustbox}{max width=0.95\textwidth}
\begin{tabular}{c c c c c c c c c c c}
\hline
\multicolumn{2}{c}{\textbf{Model}} &\textbf{MSCOCO} &\textbf{Objects365} &\textbf{Products7417}\\
 \hline
 
\multicolumn{2}{c}{YOLOv11} & 55.52	& 27.12 & 19.37\\ 
\multicolumn{2}{c}{YOLO-World} & 42.34 & 25.87 & 10.95 &  \\ 
\multicolumn{2}{c}{UniDGF (BART)} & 56.39 & 33.67 & \textbf{24.14}\\ 
\multicolumn{2}{c}{UniDGF (Pythia)} & \textbf{56.53}	& \textbf{33.78} & 13.54\\ 
\hline
\end{tabular}
\end{adjustbox}
\label{table_MAP}
\end{table}

\subsection{Overall Architecture}
Our unified framework consists of two major components: object vision embedding and generative prediction.

\textbf{Object Vision Embedding.}
As shown in Fig.~\ref{fig:fig_main}, we use a pre-trained vision backbone and keep it frozen during training. 
Each input image is processed once to produce a global feature map $F_{img}$ that retains rich spatial visual semantics.

Building on the global feature map, the pipeline in Fig.~\ref{fig:fig_main}(b) uses ground-truth bounding boxes $[x,y,w,h]$ to extract region-specific features from $F_{img}$ via ROI Align~\cite{he2017mask}, yielding the object-level
representation $F_{obj}$. A Q-Former~\cite{zhang2024vision} then attends to $F_{obj}$ with a set of learnable queries to obtain a fixed-length object representation, which is subsequently passed to the generative model for category prediction and attribute recognition.

During inference (Fig.~\ref{fig:fig_main}(c)), YOLOv11 predicts bounding boxes of objects, followed by NMS to filter redundant detections. Each remaining box is passed through ROI Align and the Q-Former to derive object embedding $E_{obj}$, which the generative model then uses to predict the corresponding categories and attributes.

\textbf{Generative Prediction.}
Hierarchical encoding naturally aligns with the next-token prediction (NTP) paradigm, since both operate on structured token dependencies. The generative model conditions on the object embedding $E_{obj}$ and autoregressively generates the hierarchical encoding of each object's category and attributes. As shown in Fig.~\ref{fig:fig_hie}, each label is encoded as a sequence of <$3,1,1$> tokens: three category tokens, followed by one attribute-property and one attribute-value token. This compact vocabulary is more efficient than standard natural-language tokenizers. 

At inference time, illustrated in Fig.~\ref{fig:fig_main}(c), the model generates desired tokens via beam search, capturing the one-to-many mapping between objects and attributes. 
Additionally, UniDGF supports property-conditioned prediction: after generating category tokens, we append a property token and decode only the corresponding attribute value, enabling generation of user-specified attributes.

\section{Experiments}
\subsection{Datasets}
We evaluate UniDGF on both open-source datasets, MSCOCO~\cite{lin2014microsoft,patterson2016coco}, Objects365~\cite{shao2019objects365}, and our constructed e-commerce dataset Products7417 with extensive category and attribute annotations. 

\textbf{MSCOCO.} MSCOCO~\cite{lin2014microsoft} is one of the most commonly used datasets in visual tasks, with annotations of 80 object categories and bounding boxes. COCO Attribute~\cite{patterson2016coco} adds 196 object attribute labels. Furthermore, we encode 80 categories and 196 attributes according to the hierarchical semantic encoding rules. Each category is mapped to token spaces in the codebook with sizes 9, 4, and 10, while each attribute is mapped to token spaces in the codebook with sizes 11 and 56. 

\textbf{Objects365.} 
Objects365~\cite{shao2019objects365} is a large-scale, high-quality dataset, containing 365 object categories. Based on these 365 categories, we further construct a three-level hierarchical category tree by GPT4. Additionally, a total 645 attributes are generated based on Qwen2.5-VL~\cite{bai2025qwen2}. 
Also, we encode the categories hierarchically into three quantized tokens with sizes 13, 5, and 26, and the attributes into two quantized tokens with sizes 44 and 71.

\textbf{Products7417.}
We collect 1 million training samples and 50,000 testing samples from the Kuaishou search platform to construct Products7417, where each sample consists of an image containing a single e-commerce product, its category label, and 3 attribute labels. The dataset covers 7,417 categories and 228 attribute values. 
The categories are hierarchically encoded into three quantized tokens with sizes 35, 43, and 115, while the attributes are encoded into two quantized tokens with sizes 4 and 173.

\subsection{Implement Details and Evaluation Metrics}
We utilize the pre-trained both CNN-based YOLOv11~\cite{khanam2024yolov11} and ViT-based DINOv3 as vision backbones, while BART~\cite{lewis2019bart} and Pythia~\cite{biderman2023pythia} as generative models. We adopt the Adam optimizer with a learning rate
of 1e-4 and a weight decay of 0.01. Additionally, we freeze the vision backbone and train the generative model for 50 epochs on 8 NVIDIA H800 GPUs.

We also propose two evaluation protocols, as shown in Table~\ref{table_ACC} and Table~\ref{table_MAP}. Table~\ref{table_ACC} provides ground truth boxes and focuses on the prediction performance on object category (<$a_{i}$>,<$b_{j}$>,<$c_{k}$>)and attribute-value (<$v_{n}$>)  with provided attribute-property (<$p_{m}$>). Table~\ref{table_MAP} evaluates object detection performance via $mAP$ with IOU threshold 0.85, as a higher IoU threshold for $mAP$ helps select high-quality bounding box predictions~\cite{Cai_2018_CVPR}.

\subsection{Experimental Results}
We evaluate UniDGF on MSCOCO, Objects365, and Products7417. As shown in Table~\ref{table_ACC} and Table~\ref{table_MAP}, UniDGF consistently improves semantic prediction and end-to-end detection quality over embedding-retrieval models, prompt-based multimodal LLMs, and classifier-based baselines. The gains are most pronounced in large-vocabulary and attribute-rich settings, demonstrating the effectiveness of our detection-driven generative paradigm.

\begin{figure}[]
    \vspace{-0.2em}
    \centering
    \begin{subfigure}[t]{0.222\textwidth}
        \centering
        \includegraphics[width=\textwidth]{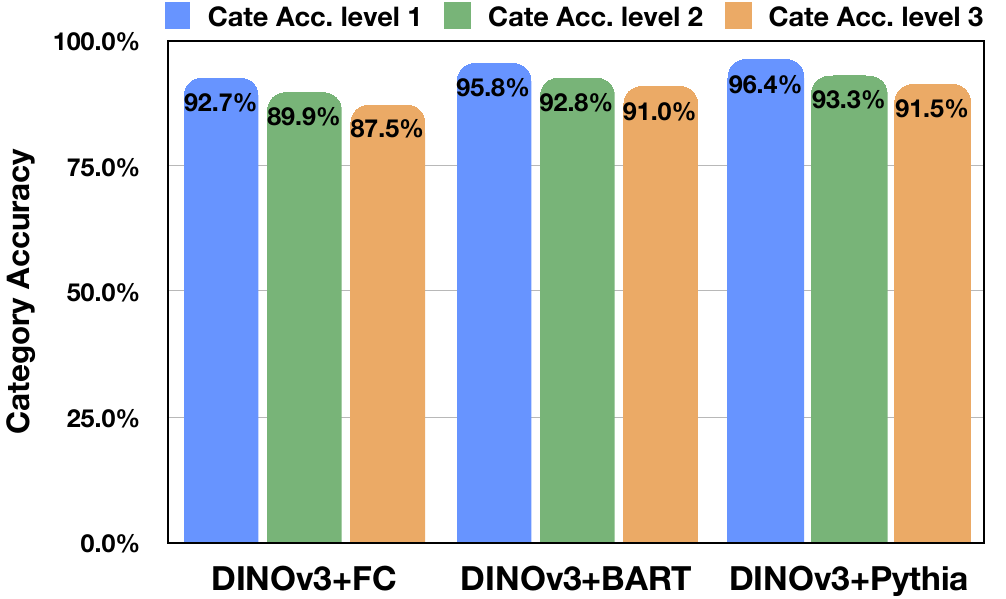}
        \caption{}
        \label{fig:subfig1}
    \end{subfigure}
    \hfill
    \begin{subfigure}[t]{0.222\textwidth}
        \centering
        \includegraphics[width=\textwidth]{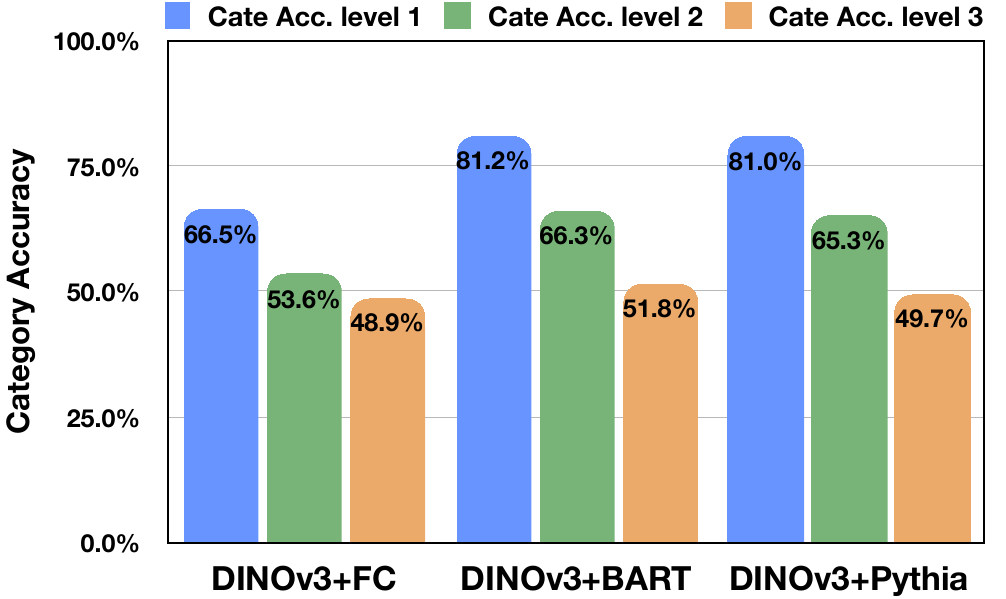}
        \caption{}
        \label{fig:subfig2}
    \end{subfigure}
    \vspace{-0.4cm}
    \caption{Comparison of category accuracy at different hierarchy levels: (a) \textbf{MSCOCO dataset}. (b) \textbf{Products7417 dataset}.}
    \label{fig:cate_level}
    \vspace{-0.2cm}
\end{figure}

\textbf{Category Prediction / Attribute Recognition (Table~\ref{table_ACC}).}
With ground-truth boxes, UniDGF shows consistent improvements across both backbones.
With YOLO backbone, UniDGF (with BART)'s encoder-decoder generation variant significantly boosts category accuracy (+13.04\% on MSCOCO, and +14.47\% on Objects365) while maintaining competitive attribute accuracy (23.35\%). On the challenging Products7417 dataset, it achieves 61.39\% category and 31.00\% attribute accuracy, outperforming embedding-retrieval methods (CLIP L/14, GME) and prompt-based models (Qwen2.5-VL-3B), which struggle with large vocabularies.
With ViT backbone, UniDGF with BART outperforms the classifier-based baseline: category accuracy +3.52\% on MSCOCO, and attribute accuracy +7.50\% on Objects365. 
The decoder-only variant (UniDGF with Pythia) shows the highest category accuracy but consistently underperforms on attribute recognition, suggesting that attribute reasoning benefits more from bidirectional encoding.
Furthermore, we evaluate the impact of different inference strategies on hierarchical category prediction accuracy. As Fig.~\ref{fig:cate_level} shows, both BART and Pythia demonstrate superior performance in hierarchical category prediction accuracy.

\textbf{End-to-End Performance (Table~\ref{table_MAP}).}
Without ground-truth boxes, UniDGF enhances end-to-end detection performance and subsequently improves downstream semantic prediction.
When paired with YOLO, UniDGF (with BART) attains higher mAP across all datasets, with the gains primarily attributed to more accurate category predictions for detected objects.
The Pythia variant achieves comparable $mAP$ on MSCOCO and Objects365.

Across three datasets and two backbones, UniDGF provides consistent improvements in both category prediction and attribute recognition. The encoder-decoder generation variant yields the best overall balance, while the decoder-only variant excels mainly in category discrimination. These results demonstrate that UniDGF forms an effective unified framework for large-vocabulary object detection and fine-grained semantic understanding.

\begin{table}[h!] 
\centering 
\caption{Ablation study on ROI align feature size and the number of Q-Former output tokens using MSCOCO~\cite{lin2014microsoft}.} 
\vspace{-0.21cm}
\label{table2}
\begin{tabular}{@{}>{\centering\arraybackslash}p{2.2cm}cc@{}}
\hline 
\textbf{ROI Output Size} & \textbf{Q-Former Output Tokens} & \textbf{Cate Acc.} \\
\hline 
1$\times$1 & 128 & 90.71 \\
5$\times$5 & 128 & 93.97 \\
7$\times$7 & 128 & \textbf{95.22} \\
9$\times$9 & 128 & 95.18 \\ 
15$\times$15 & 128 & 95.07 \\ 
7$\times$7 & 64 & 95.09 \\ 
7$\times$7 & 256 & 95.19 \\ 
\hline 
\end{tabular} 
\end{table}

\subsection{Ablation Study}
To understand the design choices in our unified detection–generation framework, we perform ablations on two factors that control the visual and semantic granularity: the ROI Align feature size and the number of Q-Former output tokens.

\textbf{ROI Align Feature Size.} 
With the token count fixed at 128, we vary the ROI spatial resolution. Extremely coarse features (1$\times$1) noticeably reduce accuracy, while performance improves as the size increases. A 7$\times$7 ROI yields the best result (95.22\%), and larger sizes (9$\times$9 or 15$\times$15) bring no further benefit, suggesting that excessive spatial detail becomes redundant.

\textbf{Number of Input Tokens.} 
Under the optimal 7$\times$7 ROI, we adjust the number of Q-Former tokens. Using only 64 tokens slightly hurts performance due to limited semantic capacity, while increasing to 256 offers only marginal gains over 128, indicating diminishing returns once token capacity is sufficient.

Overall, a 7$\times$7 ROI with 128 tokens setting provides the best trade-off between effectiveness and efficiency for hierarchical generation.

\section{Conclusion}
In this paper, we introduce a unified detection-driven generative framework that performs coherent coarse-to-fine semantic understanding for all objects in an image. 
Our method enables integrated reasoning over categories and diverse property–value pairs, outperforming classifiers, matching-based systems, and multimodal LLMs on large-scale e-commerce and open-source datasets.
In future work, we plan to extend the generative paradigm to the detection stage itself, transforming object localization and recognition into a fully unified end-to-end generative framework.

\bibliographystyle{unsrt}
\bibliography{sample-base}

\appendix

\end{document}